\pdfoutput=1
                                                           
\documentclass[a4paper, 10pt, conference]{ieeeconf}      

\newcommand{\etal}{\textit{et al.}}
\usepackage{graphicx}
\usepackage{amsmath}
\usepackage{amssymb}
\usepackage{algorithm}
\usepackage{algorithmic}
\usepackage{cite}
\usepackage[center]{caption} 
\usepackage{url} 

\IEEEoverridecommandlockouts                              
\title{\LARGE \bf
B-SDA: Bird’s-eye View Social Distancing Analysis System
}

\author{\parbox{16cm}{\centering
    {\large Zhengye~Yang$^\dagger$, Mingfei~Sun$^\dagger$, Hongzhe~Ye$^\dagger$, Zihao~Xiong$^\dagger$, Gil~Zussman$^\dagger$, Zoran~Kostic$^\dagger$}\\
    {\normalsize
    $\dagger$Department of Electrical Engineering, Columbia University, United States of America\\}}
}

\begin{document}

\maketitle

\begin{abstract}

Social distancing can reduce the infection rates in respiratory pandemics such as COVID-19. 
Traffic intersections are particularly suitable for monitoring and evaluation of social distancing behavior in metropolises. We propose and evaluate a privacy-preserving social distancing analysis system (B-SDA), which uses bird's-eye view video recordings of pedestrians who cross traffic intersections. 
We devise algorithms for video pre-processing, object detection and tracking which are rooted in the known computer-vision and deep learning techniques, but modified to address the problem of detecting very small objects/pedestrians captured by a highly elevated camera. We propose a method for incorporating pedestrian grouping for detection of social distancing violations.
B-SDA is used to compare pedestrian behavior based on pre-pandemic and pandemic videos in a major metropolitan area.
The accomplished pedestrian detection performance is $63.0\%$ $AP_{50}$ and the tracking performance is $47.6\%$ MOTA. 
The social distancing violation rate of $15.6\%$ during the pandemic is notably lower than $31.4\%$ pre-pandemic baseline, indicating that pedestrians followed CDC-prescribed social distancing recommendations. The proposed system is suitable for deployment in real-world applications.

\end{abstract}

\section{INTRODUCTION}

Respiratory viruses such as COVID-19 are spread by individuals who are in close proximity to each other for some critical period of time. Per CDC policies, individuals should maintain a social distance of at least 6 feet to suppress the spread of the virus \cite{Proof_COVID,CDC-social,CDC}. Streets and traffic intersections are locations where such social-distancing violations are prone to occur. It is desirable to provide precise measurement of social distancing in city scenarios with a possible use case of providing data that could aid epidemiological research. Furthermore, it is important to develop this technology with privacy preservation in mind.

Traffic intersections in metropolises are suitable for deployment of smart-city sensors, high-speed communications and edge computing nodes, which makes it possible to collect, process and analyze high-bandwidth data such as videos used for monitoring the social distancing behavior. The results of this paper are based on the experiments performed on a test-bed deployed in a major metropolis, which contains sensor, communications and computing infrastructure in support of privacy preserving social distancing analysis\footnote{This work was supported in part by National Science Foundation (NSF) under grant CNS-1827923.}.

In conventional traffic intersection applications, it is most common to deploy cameras at low altitudes. This approach has several drawbacks:

\begin{itemize}
\setlength\itemsep{0em}
\item The surveillance area is limited, as dictated by the directionality of low-altitude cameras.
\item Visual occlusions occur frequently for objects which are positioned behind objects closest to a camera. 
\item Temporal tracking of pedestrians is challenging due to occlusions.
\item Pedestrian face recognition and license plate recognition are possible, both of which violate personal privacy.
\end{itemize}

\begin{figure}[!t] 
\begin{center}
\includegraphics[width=0.99\linewidth]{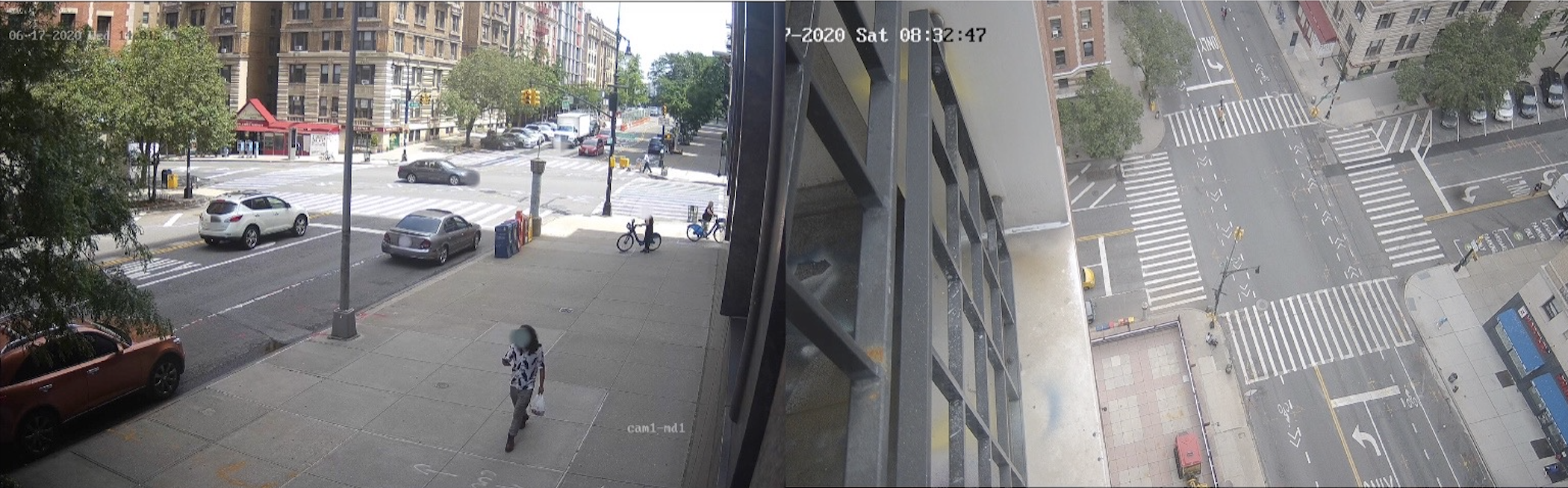}
\end{center}
\caption{Left: View from a low-elevation camera; Right: bird's-eye view from a high-elevation camera.}
\label{camera_view}
\vspace*{-0.45cm}
\end{figure}

These issues can be resolved by using cameras installed at high elevations, which record bird's-eye view videos as in  Fig.~\ref{camera_view} (right). To facilitate successful measurement of social distancing using bird's-eye camera recordings, two preliminary steps are required: (i) per-frame detection of pedestrians within the scene~\cite{7112511,Girshick_2015_ICCV,2015arXiv150601497R,2017arXiv170306870H,liu2016ssd,redmon2016you,2016arXiv161208242R,2018arXiv180402767R,2020arXiv200410934B}; (ii) reliable tracking of pedestrian trajectories across video frames~\cite{2014arXiv1409.7618L,2019arXiv190305625B,wojke2017simple,2018arXiv181011780S}. The size of pedestrians in bird's-eye view videos is a function of video resolution, and can be smaller than $30\times30$ pixels for $1080p$ recordings. Processing such small objects is a challenge for conventional object-detection and tracking algorithms. For this research, we acquired a large bird's-eye view video dataset, and annotated it for detection of pedestrians. We customized the object detection method YOLOv4~\cite{2020arXiv200410934B} and the multiple object tracking method SORT~\cite{Bewley_2016}, such that they are able to provide the desirable inference speed and to achieve promising detection and tracking performance for social distancing applications.

\begin{figure*}[!t]
\begin{center}
\includegraphics[width=0.99\linewidth]{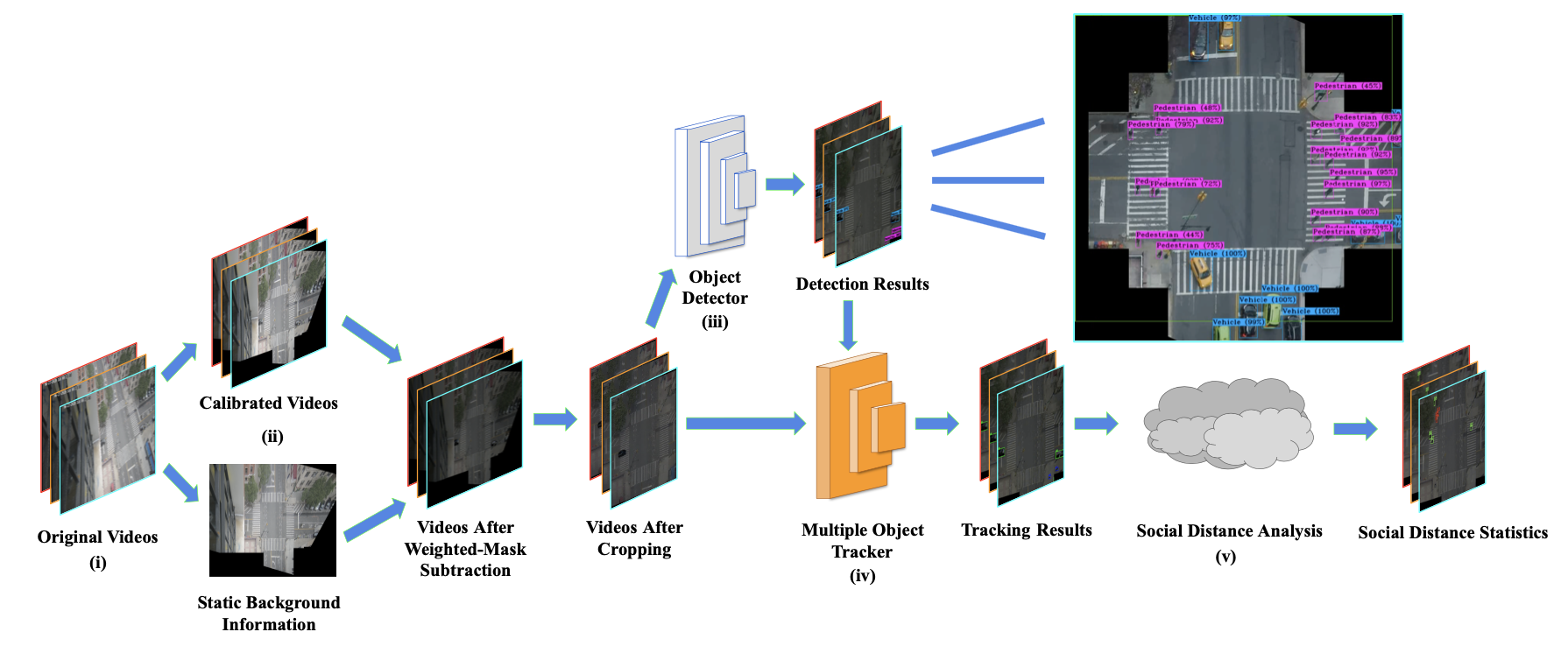}
\vspace*{-0.3cm}
\end{center}
\caption{Pipeline for the B-SDA system: (i) collect raw videos from a bird's-eye view camera; (ii) do calibration and background subtraction to alleviate the effect of sub-optimal sensor quality; (iii) get pedestrian detection results; (iv) get pedestrian tracking information; (v) analyze pedestrian movement behaviour with social distancing analysis algorithm.}
\label{system_overview}
\vspace*{-0.3cm}
\end{figure*}

A number of social distancing analysis prototypes have been recently proposed \cite{deepak,LandingAI,2020arXiv200501385S,2020arXiv200703578Y}. They are based on open-source datasets with low-altitude camera views, leading to potential privacy violations. Most utilize only the detection information to analyze social distancing, which is insufficient for acquisition of statistics such as pedestrian throughput at traffic intersections. When using only the distance between pedestrians to assess proximity violations, it is impossible to discriminate between "safe" social groups and random people in close proximity to each other. A safe social group is defined as a collection of people  assumed to reside together, such as a family. Identifying "safe" social groups requires group tracking, in order not to declare the participating pedestrians as violators of social distancing.  

The main contributions of this paper are:

\begin{itemize}
    \setlength\itemsep{0em}
    \item Social distancing analysis system based on bird's-eye view videos (Fig.~\ref{system_overview}).
    \item Introduction of a vision-based social group identity validation algorithm, for bird's-eye view based social distancing analysis.
    \item Analysis of social distancing behavior in a major metropolis, based on privacy-preserving videos recorded before and during the COVID-19 pandemic. 
\end{itemize}

\section{Background}

\subsection{Object Detection}

Object detection is a computer-vision task that classifies and locates instances of semantic objects. Most state-of-the-art object detectors are deep-learning based. Among the prominent approaches, R-CNN~\cite{7112511}, Fast R-CNN~\cite{Girshick_2015_ICCV}, Faster R-CNN~\cite{2015arXiv150601497R} and Mask R-CNN~\cite{2017arXiv170306870H} use a two-stage structure for object detection, which separates region proposal stage and classification stage. 

By contrast, SSD~\cite{liu2016ssd} and YOLO methods ~\cite{redmon2016you,2016arXiv161208242R,2018arXiv180402767R,2020arXiv200410934B} use a single-stage structure with higher inference speed. As stronger backbone networks are designed and more data augmentation methods are applied, single-stage models evolved into many variations. Recent improvements of YOLO ~\cite{2020arXiv200410934B,2016arXiv161208242R,2018arXiv180402767R,2019arXiv191108287Z} make it possible to approach the detection accuracy comparable to R-CNN's accuracy without sacrificing YOLO's inference speed in our traffic scenario. Considering the speed-accuracy trade-off, we chose YOLOv4~\cite{2020arXiv200410934B} as the baseline for object detection in this work.

\subsection{Multiple Object Tracking}

The task of Multiple Object Tracking (MOT) is largely partitioned into locating multiple objects, maintaining their identities, and yielding the individual trajectories given an input video~\cite{2014arXiv1409.7618L}. 

SORT~\cite{Bewley_2016} is a tracking-by-detection algorithm in which the aim is to associate detection across frames in a video sequence. It  performs imperfectly when tracking through occlusions. DeepSORT~\cite{wojke2017simple} algorithm replaces the association metric with a more informed metric that combines motion and appearance information by using a convolutional neural network. Tracktor method~\cite{2019arXiv190305625B} accomplishes multi-object tracking with the Faster R-CNN object detector only, but it under-performs when the objects are moving fast or when the density of objects is high.

Given the similar performances of these methods tested in our scenario, we chose SORT~\cite{Bewley_2016} as the tracking module due to its simplicity and quick inference.

\subsection{Social Distancing Analysis}

The general aspects of the visual social distancing problem have been proposed in \cite{cristani2020visual}. With the sudden outbreak of the COVID pandemic, the proposals for social distancing surveillance have been rapidly emerging. 

Given the nature of the social distancing problem, the primal task is to detect pedestrians and measure the distance between each pair. In early works, the solutions were based on object detection with distance approximation\cite{deepak,LandingAI,2020arXiv200703578Y}.  Current mainstream object detection frameworks such as Faster R-CNN\cite{2015arXiv150601497R}, Mask R-CNN~\cite{2017arXiv170306870H}, YOLOv3~\cite{2018arXiv180402767R} and YOLOv4~\cite{2020arXiv200410934B} are widely used. To collect more useful identity-related statistics, tracking-by-detection MOT method has been added into the workflow by \cite{punn2020monitoring,gupta2020sd}. Considering the accuracy-speed tradeoff in tracking algorithms, efficient methods like SORT and DeepSORT are widely used for social distancing problems. Group information among crowds is further added to reduce false positives caused by naïvely chosen distance thresholds, and several works proposed different group detection mechanisms\cite{rezaei2020deepsocial,sun2020autonomous}.

All research noted above is based on street-level cameras and dataset \cite{Oxford}, which violates pedestrian privacy. Bird's-eye view cameras provide an alternative which achieves privacy preservation and provides a much larger surveillance area per camera.

\subsection{Group Detection}

Due to the inherent social nature of human behavior, interactions typically happen between small subsets of people referred to as groups~\cite{6636608}. Hall proposed the proxemic theory which defined distances between two people with different intimacy levels in the North American culture~\cite{Hall1966Hidden}. Many researchers investigated measurement metrics for group identification - Rima \etal proposed similarity features including velocity and tracklets~\cite{10.1016/j.patcog.2016.06.016}; Francesco \etal conducted proxemics and causality for group identification~\cite{6636608}; Weina \etal proposed new measurements considering trajectory and velocity comparisons, and constructed a bottom-up hierarchical clustering approach which starts from separate individuals and gradually build larger groups by merging clusters with strongest intergroup closeness~\cite{10.1109/TPAMI.2011.176}. Mahsa \etal proposed an end-to-end trainable framework that are able to generate spatio-temporal representation from and extract features of each person and learn the representation of collective activities~\cite{2020arXiv200702632E}.

Crowd understanding, or crowd analysis, a topic related to group detection, is also an active research field. Ning \etal proposed an attention-injective deformable convolutional network called ADCrowdNet, which could address the accuracy degradation problem of highly congested noisy scenes~\cite{2018arXiv181111968L}. Yuting \etal developed a network that can handle both detection and crowded counting without annotation with bounding boxes~\cite{8953871}.

\section{Description of the B-SDA System for Measuring and Analysis of Social Distancing}

\subsection{Data Pre-Processing}

The work described in this paper is based on bird's-eye view video recordings.
The use of highly elevated cameras, desirable for privacy protection reasons, results in small and potentially blurry pedestrians. Video recordings under various lighting and weather conditions additionally impact the accuracy of object detection and tracking. To tackle these challenges, we apply data pre-processing methods - Weighted-Mask Background Subtraction (WMBS) and Video Calibration (VC). 

WMBS constructs the background image from videos acquired by static cameras, computed as the mean of all $N$ frames~\cite{VideoAveraging}. 


The background image, with a weighted parameter $\alpha$,  is subtracted from the original frames, to obtain the enhanced video shown in Eq.~\eqref{back_ground_for}:

\begin{equation}\label{back_ground_for}
    F_b(I_{r}^{(t)})=I_{r}^{(t)} - \frac{\alpha}{N}\sum_{k=1}^{N}I_{r}^{(k)}.
\end{equation}

$I_b^{(t)}=F_b(I_r^{(t)})$ represents the output image where $I_{r}^{(t)}$ is $t$-th frame in the original video, and $\alpha$ is the weight coefficient. 

The benefits of WMBS are:
\begin{itemize}
    \item Sharpening of the boundaries between objects and the static background.
    \item Minimization of the influence of environmental factors including brightness, contrast and weather conditions, while allowing the detection of static objects. 
    \item Improvement in detection performance, with low additional computational complexity.
\end{itemize}

VC transforms original (slanted) bird's-eye view videos into true/calibrated bird's-eye videos that are perpendicular to the ground. It maps a trapezoidally distorted traffic intersection scene into a rectangular one with a uniform scale, and is performed prior to object detection and tracking. It facilitates accurate localization of pedestrians while presenting all objects "as seen straight from above". Calibration is achieved by calculating the homography matrix $M_{ca}$ that maps $I_{b}^{(t)}$ in image coordinates to $F_c(I_{b}^{(t)})$ in real world coordinates. The calibration formula is shown in Eq.~\eqref{calibration}. 

\begin{equation}\label{calibration}
    F_{ca}(I_{b}^{(t)})=M_{ca} \cdot I_{b}^{(t)}.
\end{equation}

Center cropping is the final stage in calibration, which removes parts of the original image without useful content and increases the per-pixel size of features,  to improve the detection performance. The cropped image $I^{(t)}$ is the input for procedures that follow.


\subsection{Detection and Tracking}

The goal of object detection in tracking-by-detection is to provide object localization and classification information to a Multiple Object Tracking (MOT) algorithm to associate object identities. The detector provides $m^{(t)}$ proposed bounding boxes $D^{(t)}=\{d_1^{(t)},d_2^{(t)},...,d_{m^{(t)}}^{(t)}\}$ from $I^{(t)}$, and $d_j^{(t)}=(dclass_j^{(t)},bbox_j^{(t)},conf_j^{(t)})$ records the class, location and confidence score of the $j$-th predicted box. 

Once the MOT algorithm receives $D^{(t)}$, identity association is performed to get the tracking state $S^{(t)}=\{s_1^{(t)},s_2^{(t)},...,s_{m^{(t)}}^{(t)}\}$ for all $m^{(t)}$ objects in the $t$-th frame. $s_j^{(t)}=(class_j^{(t)},roi_j^{(t)}, id_j^{(t)})$ denotes the state of the $j$-th object in the $t$-th frame, $class_j^{(t)}$ is the class of that object, $roi_j^{(t)}$ is the location of that object, and $id_j^{(t)}$ indicates the unique ID for that object. 

\subsection{Social distancing analysis}

\begin{algorithm}[t]
\caption{Group Validation}
\label{alg:grp_validation}
\begin{algorithmic} 
\REQUIRE $\textrm{Object state history: }T^{(t)} ,\textrm{Violation list: } L^{(t)}$
\STATE \textbf{For} {violation pair$(i,j)$ in $L^{(t)}$:  }
\STATE \:\:\:\:$ obj_{i}^{(t)} \leftarrow T^{(t)}[i]$
\STATE \:\:\:\:$ obj_{j}^{(t)} \leftarrow T^{(t)}[j]$
\STATE \textbf{If} {Velocity Compare($obj_{i}^{(t)}, obj_{(j)}^{t}$) = \textbf{True}}
\STATE \:\:\:\:\textbf{If} {Trajectory Compare($obj_{i}^{(t)}, obj_{j}^{(t)}$) = \textbf{True}}
\STATE \:\:\:\:\:\:\:\:$\textrm{remove }L_t[(i,j)]$
\RETURN $L_t$
\end{algorithmic}
\end{algorithm}

The social distancing analysis system continually receives the tracking information $S^{(t)}$ for each frame. The system keeps updating the whole tracking state $S$ and extracts useful information for analysis from $S$ to create the state history $T = \{T^{(1)},T^{(2)},...,T^{(t)}\}$. Here, $T^{(t)} = \{obj_1^{(t)},obj_2^{(t)},...,obj_{m^{(t)}}^{(t)}\}$ and $obj_{j}^{(t)} = (roi_{j}^{(t)},bp_{j}^{(t)},id_{j}^{(t)},v_{j}^{(t)},traj_{j}^{(t)})$ denote the state of $j$-th object in the $t$-th frame. $roi_{j}^{(t)}$ and $id_{j}^{(t)}$ are received directly from $s_j^{(t)}$ in $S^{(t)}$ which represents the object bounding box information and the unique ID for this object. $bp_{j}^{(t)}$ represents the middle point of the bounding box bottom which can be derived from $roi_{j}^{(t)}$. $traj_{j}^{(t)}$ records all previous tracking information for this object.

The estimation of real-world distances between objects is simplified by the bird's-eye video calibration. The distance of six feet in our videos is represented by approximately 35 pixels based on our ground measurement. For original uncalibrated images, there is a significant difference in scale distortion between X and Y axes. This difference becomes negligible after  calibration has been applied.

Next we create an upper triangular Euclidean distance matrix for all objects in $T^{(t)}$ to find potential social distancing violation pairs $L^{(t)}=\{(bid^{(t)}_{11},bid^{(t)}_{12}),...,(bid^{(t)}_{p1},bid^{(t)}_{p2})\}$, where $\{bid^{(t)}_{n1},bid^{(t)}_{n2}\}$ denotes the $n$-th violation object ID pairs of all $p$ pairs. This is the final violation criteria for most contemporary social distancing analysis systems~\cite{deepak,LandingAI,2020arXiv200501385S,2020arXiv200703578Y}, although there may still be a number of pairs in $L^{(t)}$ that belong to the same group. 

In order to avoid over-counting the number of violations, we design a cascade-condition filter shown in Algorithm~\ref{alg:grp_validation} to validate if a social distancing violation pair is (incorrectly) indicated, although the pedestrians belong to the same social group. The assumption made here is that people belonging to the same (safe) social group are going to maintain the proxemic relationship when they cross the traffic intersection. The proxemic relationship can be captured by pedestrian trajectory information, which can be decomposed into three components: velocity similarity, trajectory similarity\cite{magdy2015review} and proxemic stability. 

Trajectory similarity is measured by the Euclidean distance between two pedestrians at the same temporal location shown in Eq.~\eqref{traj_similarity}. The first order derivative of a trajectory can be used as velocity estimator.

\begin{equation}\label{traj_similarity}
    \begin{aligned}
    sim(a,b)&=\frac{\sum_{i \in X_{a}\bigcap X_{b}} \|(a,b)\|_{2}}{|X_{a}\bigcap X_{b}|}
    \end{aligned}
\end{equation}

The estimation of object velocity depends heavily on the correct localization of the bounding boxes. Caused by imperfections of the object detection algorithm, oscillations in localization could seriously disturb the estimation of object velocities. To achieve the precise velocity comparison between two objects, we calculate velocities with Exponentially Weighted Averages shown in Algorithm~\ref{alg:Velocity_update}. Function \textbf{Get Bp} computes the middle point of the bounding box bottom in frames, and function \textbf{Find} collects frames where a given object is detected. When exponentially weighted averaging is applied, velocities are recalculated based on previous and current velocities using the update parameter $\alpha$.

Assume that there are two pedestrians with velocity vectors $v_1$ and $v_2$. To compare the velocity similarity, both magnitude and direction need to be evaluated. The Cosine Distance ($D_{cos}$) Eq.~\eqref{cos_similarity} is a common method to measure the similarity in vector direction.

\begin{equation}\label{cos_similarity}
 D_{cos} = 1-  \frac{v_{1}\cdot v_{2}}{\parallel v_{1} \parallel \cdot \parallel v_{2} \parallel }. \\
\end{equation}

In order to combine the magnitude similarity with the Cosine Distance measure, it needs to be scaled. We use the magnitude similarity ($D_{Mag}$) which is inspired by the formula discussed by Rima~\etal~\cite{10.1016/j.patcog.2016.06.016}

\begin{equation}\label{norm_euc} 
 D_{Mag} = \frac{|\parallel v_1\parallel-\parallel v_2\parallel|}{\max{(\parallel v_1\parallel, \parallel v_2\parallel)}}. \\
\end{equation}

The overall velocity distance $D(v1,v2)$ with weight parameter $\gamma$ is described in Eq.~\eqref{velocity_similarity}

\begin{equation}\label{velocity_similarity}
 D(v1,v2) = \gamma\cdot D_{cos} + (1-\gamma)\cdot D_{Mag}. \\
\end{equation}

The ratio between standard deviation of distance and trajectory similarity between two violation candidates captures the stability of the proxemic relationship, shown in Eq.~\eqref{traj_stability}. The function Trajectory Compare is the "OR" condition filter between trajectory similarity and trajectory stability, which aims to be suitable for different group scenarios in a traffic intersection. 

\begin{equation}\label{traj_stability}
    \begin{aligned}
    f_{\textrm{stab}}&=\frac{std.(a,b)}{sim(a,b)}
    \end{aligned}
\end{equation}

\begin{algorithm}[tp]
\caption{Velocity update}
\label{alg:Velocity_update}
\begin{algorithmic} 

\REQUIRE $\textrm{Tracking state: }S$
\STATE \textbf{For} {$obj$ in S: }
\STATE \:\:\:\:$bp[obj]=\textrm{Get Bp}(S,obj)$
\STATE \:\:\:\:\textrm{Find} $[t_0,t_1,...,t_n]$ that $obj$ exists 
\STATE \:\:\:\:\textbf{For} {$t_{k}$ in $[t_0,t_1,...,t_n]$ }
\STATE \:\:\:\:\:\:\:\:\textbf{If} {$k == 0$:}
\STATE \:\:\:\:\:\:\:\:\:\:\:\:$v^{(t_{k})} = 0$
\STATE \:\:\:\:\:\:\:\:\textbf{Else:}
\STATE \:\:\:\:\:\:\:\:\:\:\:\:$v^{(t_{k})} = \frac{bp[obj][t_{k}]-bp[obj][t_{k-1}]}{t_{k}-t_{k-1}}$
\STATE \:\:\:\:\:\:\:\:\textbf{If} {apply Exponentially Weighted Averages:}
\STATE \:\:\:\:\:\:\:\:\:\:\:\:\textbf{If} {$k \leq 1 $:}
\STATE \:\:\:\:\:\:\:\:\:\:\:\:\:\:\:\:$v^{(t_{k})}=v^{(t_{cur})}$
\STATE \:\:\:\:\:\:\:\:\:\:\:\:\textbf{Else} {$k\leq n $:}
\STATE \:\:\:\:\:\:\:\:\:\:\:\:\:\:\:\:$v^{(t_{k})}=\alpha v^{(t_{k})} + (1-\alpha)v^{(t_{k-1})}$
\end{algorithmic}
\end{algorithm}

The function Velocity Compare calculates the velocity similarity between two objects indicated to be in a social-distancing violation. If the velocity similarity lies within the predefined threshold, these two objects will be further evaluated by the function Trajectory Compare. If all comparisons pass the test, these two objects are declared to belong to the same group, and they are removed from the social distancing violation list $L^{(t)}$. After these modifications, the list can be used as an index to collect the statistics.
 
\section{Experiments}

\subsection{Data Acquisition}

The dataset for training has been composed from two resources: (i) public dataset called Visdrone2019~\cite{2020arXiv200106303Z}, and (ii) a collection of traffic intersection videos recorded in New York City, and annotated by our group (B-SDA dataset). Our videos have been recorded by Hikvision (DS-2CD5585G0-IZHS) camera. The videos have been recorded at 15 frames per second using $1920\times 1080$ resolution. The  annotation statistics are shown in Table~\ref{table:trainingset}. For inference purposes, videos have been systematically recorded multiple times per day from June 2020 to Febuary 2021 during the pandemic, with recording schedule and statistics shown in Table \ref{table:schedule}.

\begin{table}[tp]
\caption{Annotation statistics}
\begin{center}
{
\footnotesize

\begin{tabular}{|c|c|c|}
\hline
\textbf{Dataset} & \textbf{Number of Frames} & \textbf{Number of Objects}\\
\hline\hline
B-SDA train & 7.4k & 49.7k\\
\cline{1-3}
B-SDA test & 8.1k & 203.2k\\
\cline{1-3}
\hline
\end{tabular}

}
\end{center}
\label{table:trainingset}
\vspace*{-0.5cm}
\end{table}

\begin{table}[btp]
\caption{Recording schedule and video statistics during the pandemic}
\begin{center}
{
\footnotesize

\begin{tabular}{|c|c|}
\hline
\textbf{Recording Schedule} & \textbf{Number of Frames} \\
\hline\hline
09:00-09:05 & 137.7k  \\
\cline{1-2} 
14:00-14:05 & 140.4k \\
\cline{1-2}
17:30-17:35 & 148.5k \\
\cline{1-2}
22:00-22:05 & 143.1k \\
\cline{1-2}
\hline
\end{tabular}
}
\end{center}
\vspace*{-0.5cm}
\label{table:schedule}
\end{table}


\subsection{Experimental Setup}
\label{experimental setup}
The computational setup consists of the operating system Ubuntu 18.04 running on a cluster of 8 vCPUs, 30GB RAM, and one Tesla P100 GPU.

To reach detection accuracy appropriate for the proposed social distancing analysis system, we customized YOLOv4 object detection algorithm in the following ways: (i) it is using a shallower feature map ($\frac{1}{4}$ of input image size) was chosen to detect small pedestrians; (ii) the scale factors of the anchor boxes were determined based on experimental studies to adjust them to the scale of a target class. In the training process of YOLOv4, the customized YOLOv4 started with the backbone pre-trained on the Imagenet dataset~\cite{2014arXiv1409.0575R}. Next, it is trained with (a) VisDrone2019 dataset~\cite{2020arXiv200106303Z} in $832\times832$ resolution for 6,000 epochs, followed by (b) B-SDA dataset for another 6,000 epochs. We used a batch size of $64$ and the learning rate of $10^{-3}$ with a weight decay of $5\times10^{-4}$. 

For tracking, we used the SORT algorithm~\cite{Bewley_2016} for real-time processing without sacrificing much in accuracy.

The complete group validation is shown in algorithm~\ref{alg:grp_validation}. We use Eq.~\eqref{cos_similarity}, Eq.~\eqref{norm_euc} and Eq.~\eqref{velocity_similarity} to check the velocity similarity. Parameters $\gamma$ and $\lambda$ are set to 0.1 and 1 respectively. The threshold for velocity similarity is set to 0.21.

For similarity measurement for trajectories, we use Eq.~\eqref{traj_similarity} and Eq.~\eqref{traj_stability}. The social distance threshold for Eq.~\eqref{traj_similarity} is $35$. The trajectory stability threshold for Eq.~\eqref{traj_stability} is set to $0.25$.

\subsection{Verification of the Group Validation Algorithm}
We evaluate the effectiveness of our algorithm for validation of pedestrian groups. 

We annotate groups of pedestrians with larger bounding boxes that cover all pedestrians within a same group, for $10k$ frames.

In each group bounding box, pedestrians who are within the social distancing threshold (35 pixels) are the true positives in the group validation evaluation. The corresponding metrics are shown below.
\begin{equation}\label{Precision}
    \begin{aligned}
    \textrm{Precision}=\frac{TP}{TP + FP}
    \end{aligned}
    \vspace*{-0.2cm}
\end{equation}

\begin{equation}\label{Recall}
    \begin{aligned}
    \textrm{Recall}=\frac{TP}{TP + FN}
    \end{aligned}
    \vspace*{-0.2cm}
\end{equation}

\begin{equation}\label{F1}
    \begin{aligned}
    \textrm{F1-score}=\frac{2\times \text{Precision} \times \text{Recall}}{\text{Precision} + \text{Recall}}
    \end{aligned}
\end{equation}

\begin{table}[!t]
\caption{Group Validation Performance}
\begin{center}
{
\footnotesize

\begin{tabular}{|c|c|c|c|c|c|}
\hline
\textbf{Traj. Comp.} & \textbf{Vel. Comp.} & \textbf{Precision} & \textbf{Recall} & \textbf{F1}\\
\hline\hline
 & & \textbf{0.92} & 0.57 & 0.66   \\
\cline{3-5}
\checkmark &  & 0.90 & \textbf{0.99} & \textbf{0.92}   \\
\cline{3-5}
\checkmark & \checkmark & 0.86 & 0.96 & 0.88   \\
\hline
\end{tabular}
}
\label{group_validation_performance}
\vspace*{-0.5cm}
\end{center}
\end{table}

Table~\ref{group_validation_performance} shows that our algorithm can capture accurate grouping information, and filter out violation pairs in the same group (Traj. Comp. denotes Trajectory Compare and Vel. Comp. denotes Velocity Compare). Different combination of methods are experimented with. We observe that Trajectory Comparison brings significant improvement to the Recall and to the F1 score, whereas Velocity Comparison is harmful to the performance. We postulate that velocity estimation introduces a large amount of noise even when equipped with the exponentially weighted average. Therefore, we remove the function Velocity Estimation, and use only the function Trajectory Comparison for further analysis of social distancing.

\section{Results}
\subsection{Detection and Tracking for bird's-eye view videos}

\begin{table}[!t]
\caption{YOLOv4 performance on the B-SDA Dataset with data pre-processing applied}
\begin{center}
{
\footnotesize

\begin{tabular}{|c|c|c|c|c|c|}
\hline
 \textbf{WMBS} & \textbf{CC} & \textbf{AP}& \textbf{mIoU}& \textbf{Precision} & \textbf{Recall}\\
\hline\hline
\cline{1-4}
    &  & 44.9 & \textbf{71.7} & 74.2 & 49.9   \\
   \checkmark &  & 55.1 & 69.8 & 70.9 & 62.9   \\
    &\checkmark & 58.0 & 68.75 & \textbf{84.1} & 62.8  \\
   \checkmark & \checkmark & \textbf{63.0} &68.77 & 73.3 & \textbf{73.0}   \\
\hline
\end{tabular}
}
\label{tab_yolo_result}
\vspace*{-0.5cm}
\end{center}
\end{table}

Precise object detection and tracking results are essential to our social distancing system. Table~\ref{tab_yolo_result} shows how data pre-processing methods affect our customized YOLOv4 model. We use \textbf{WMBS} for Weighted-Mask Background Subtraction and \textbf{CC} for Center Cropping. With these two methods, YOLOv4 achieves the highest AP and Recall compared to other combinations. For the traffic surveillance scenario, considering the traffic safety, recall is more important than precision. We choose the YOLOv4 model with both pre-processing methods for the detector.

The overall MOT accuracy is evaluated by the CLEAR metrics~\cite{Bernardin2008Evaluating}, where MOTA is the main performance evaluation score. The tracking performance is evaluated on the B-SDA test dataset. The detection result is generated by YOLOv4 with Weighted-Mask Background Subtraction and Center Cropping. For the YOLOv4-SORT pipeline, 
we obtain MOTA = 47.65\%, MOTP = 71.4\%, MT = 60.9\%, and ML = 5.8\% .

The YOLOv4-SORT pipeline provides accurate object locations and identity information across all frames, with which we can measure the social distances between pedestrians. Furthermore, our group validation algorithm can obtain reliable social distancing analysis.

\subsection{Social Distancing Analysis}

\begin{figure}[!t]
\includegraphics[width=0.99\linewidth]{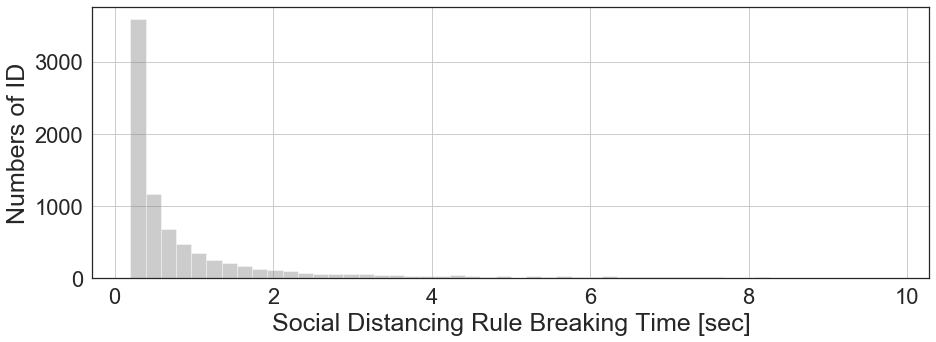}
\caption{Distribution of the duration of social distancing violations (raw statistics).}
\vspace*{-0.2cm}
\label{Social_breaking_bar_no_filter}
\end{figure}

Fig.~\ref{Social_breaking_bar_no_filter} shows the statistical information about the time duration of violations for all social distancing violators in 458 videos recorded during the COVID-19 pandemic in a major metropolis. For comparison purposes, we also performed the social distancing analysis on another B-SDA video dataset which was collected between June and July 2019, before the COVID-19 pandemic.

To simplify the evaluation and visualization of the statistics, we note that walking pedestrians may not be able to maintain the social distance of exactly 6 feet when crossing a street, which will cause violations that are less than one second in duration. These short violations would count pedestrians who behave properly as social distancing violators. Therefore, in the following analysis and figures we omit all violations which are shorter than 1 second.

The estimation of pedestrian volume (number of pedestrians) relies on the number of trajectories. ID switches caused by imperfect tracking enlarge the estimated number of pedestrians. To better estimate the pedestrian volume, the real world average trajectory length can be used as the reference. We annotated ground truth for five video segments recorded during the pandemic, from which we calculated that the real world average pedestrian trajectory length is $19.11$ seconds. In the 458 recordings, the average pedestrian trajectory length extracted from tracking inference is equal to $5.63$ seconds. This decrease in the duration of trajectories, from $19$ to $5$ seconds, reflects the fact that the ID switches enlarge the perceived number of pedestrians. To alleviate this effect, we apply \eqref{estimate_volume} to estimate the actual pedestrian volume ($EV$). $TC$ stands for the total number of trajectories, $AT_{GT}$ for the average trajectory length from the ground truth, and $AT_{Infer}$ stands for the average trajectory length obtained from inference results. 

\begin{equation}\label{estimate_volume}
\begin{aligned}
EV = TC \cdot \frac{AT_{\textrm{GT}}}{AT_{\textrm{Infer}}}
\end{aligned}
\end{equation}

\begin{figure}[!t]
\begin{center}

\includegraphics[width=0.99\linewidth]{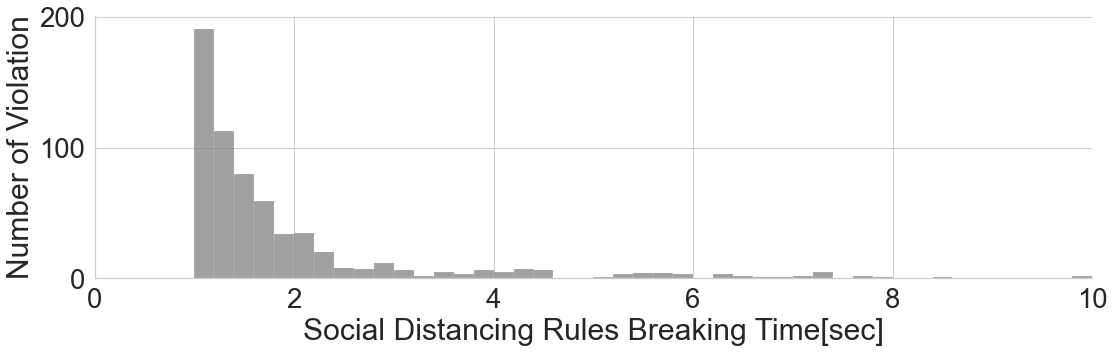}
\end{center}
\vspace*{-0.2cm}
\caption{The distribution of social distancing violation duration.}
\vspace*{-0.3cm}
\label{Social_breaking_bar}
\end{figure}

\begin{figure}[!t]
\begin{center}
\vspace*{-0.1cm}

\includegraphics[width=0.99\linewidth]{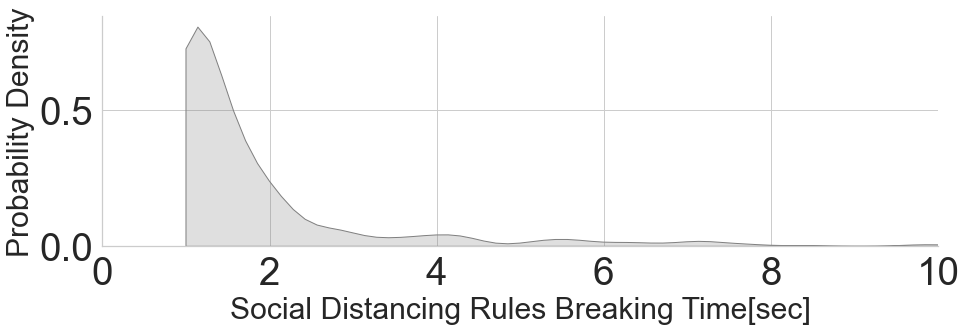}
\end{center}
\vspace*{-0.2cm}
\caption{The distribution of social distancing violation duration (estimated by KDE using Gaussian kernel with 0.08 bandwidth).}
\label{Social_breaking}
\end{figure}

Fig.~\ref{Social_breaking_bar} shows the histogram of the duration of violations, which is filtered by kernel density estimation (KDE) to obtain the distribution shown in Fig.~\ref{Social_breaking}. Over 75 percent of violations are shorter than 4 seconds.

\begin{figure}[!t]
\begin{center}

\includegraphics[width=0.99\linewidth]{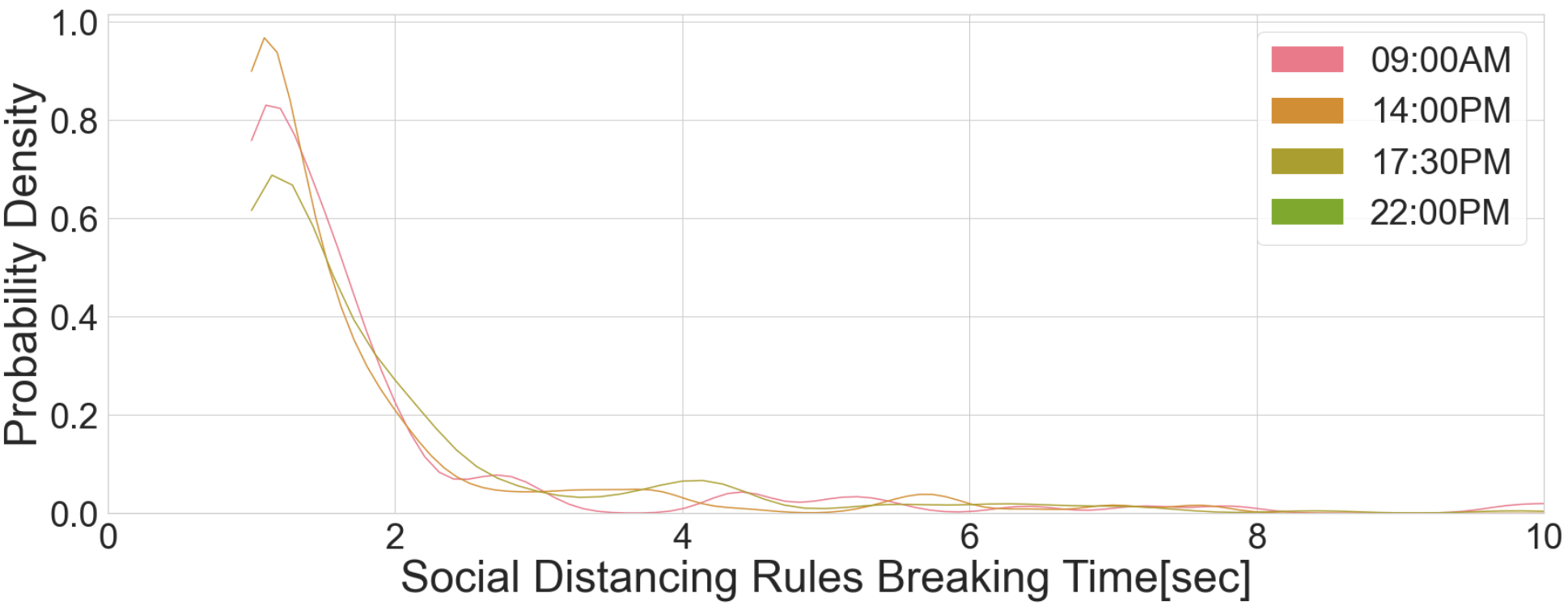}
\end{center}
\vspace*{-0.2cm}
\caption{Probability distribution for the social distancing violation duration in different times of the day (estimated by KDE with Gaussian kernel).}
\label{Social_breaking_time}
\end{figure}

\begin{figure}[!t]
\begin{center}

\includegraphics[width=0.99\linewidth]{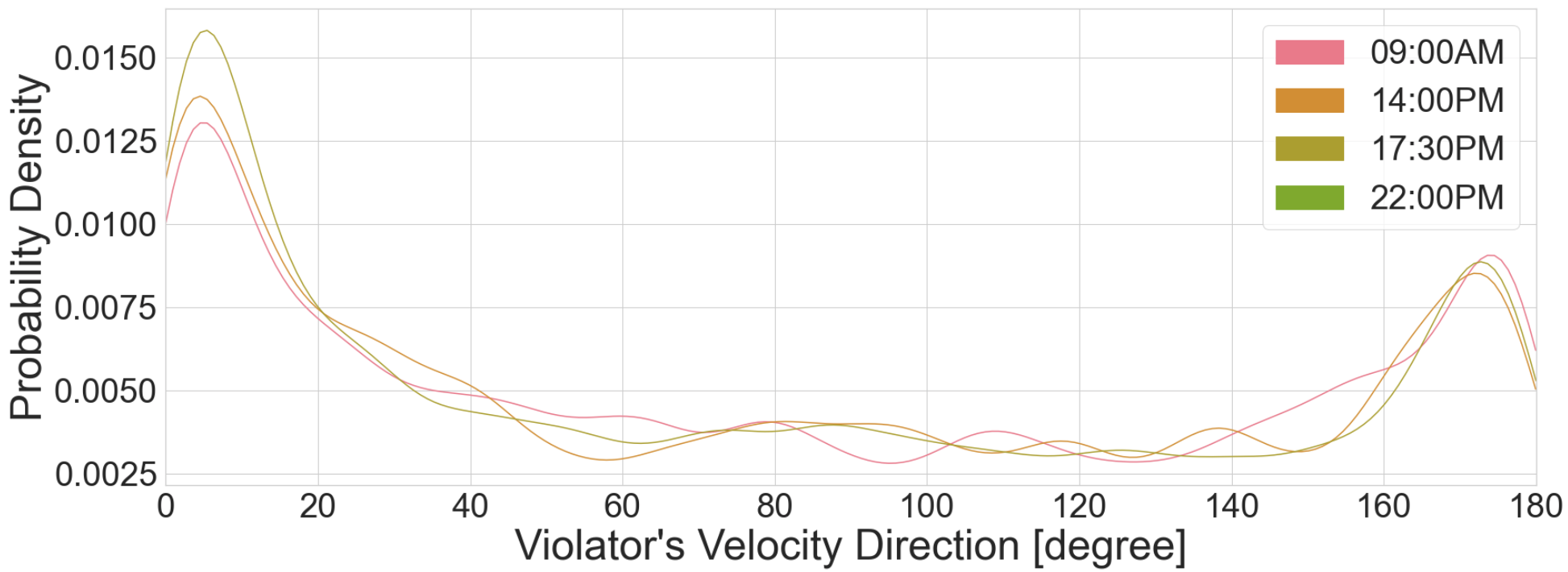}
\end{center}
\vspace*{-0.2cm}
\caption{Distributions of the angle between moving directions of two pedestrians in a violation pair at different times of day (estimated by KDE with Gaussian kernel).} 
\vspace*{-0.3cm}
\label{velocity_angle_time_slot}
\end{figure}

Since our pandemic videos were recorded daily on a repetitive time schedule, we can generate probability density distributions for each of the recording periods. These distributions smoothed by KDE are shown in Fig.~\ref{Social_breaking_time}. There is limited difference between different time slots. To check if the behavior of the pedestrians is affected by the time of the day, we calculate the velocity direction for each violation pair, and we record the angular difference in directions to better depict the violation situation. The corresponding distributions are shown in Fig.~\ref{velocity_angle_time_slot}, which again indicate that the time of day does not impact the statistics.

\begin{figure}[!t]
\begin{center}
\includegraphics[width=0.99\linewidth]{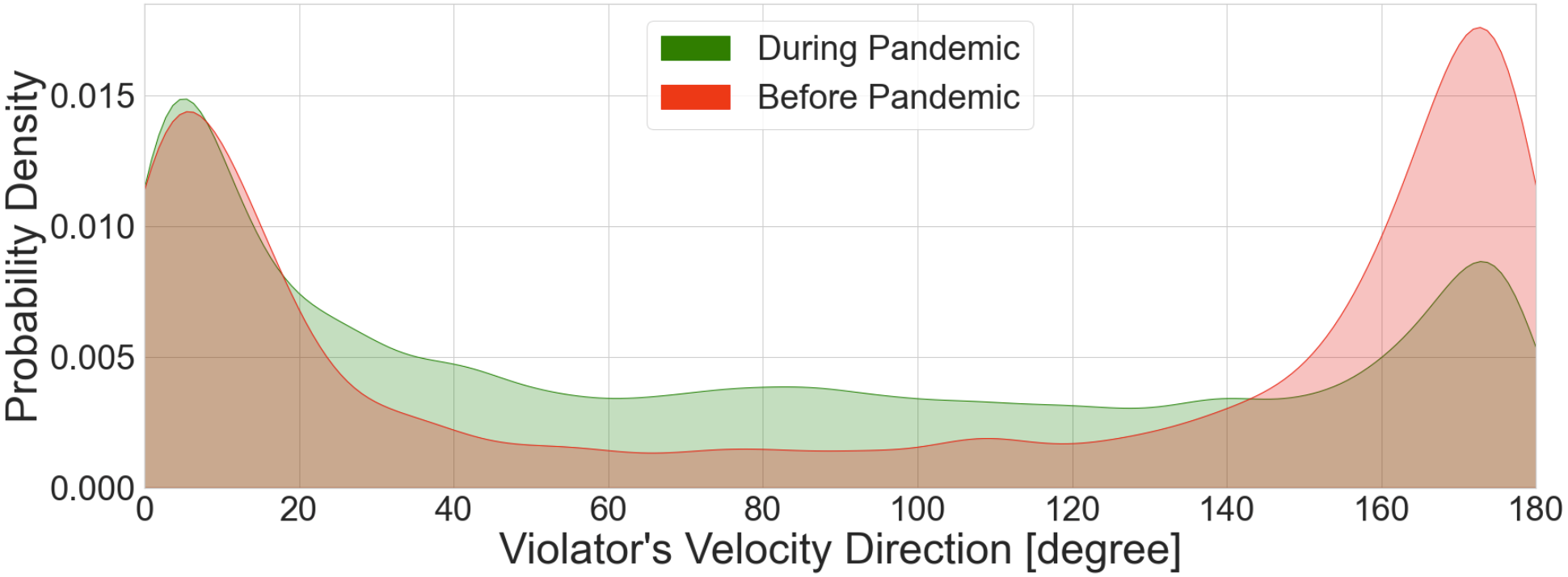}
\end{center}
\vspace*{-0.2cm}
\caption{Distribution of the angle between moving directions of two pedestrians in a violation pair (estimated by KDE with Gaussian kernel).}
\label{angle_comparison}
\vspace*{-0.5cm}
\end{figure}

Fig.~\ref{angle_comparison} shows the probability distribution of the angle between moving directions of two pedestrians in a violation pair before and during the pandemic. We define face-to-face violation happens when the difference of velocity direction is larger than 150 degrees. Before the pandemic, 42.3\% of violations are face-to-face. During the pandemic, the distribution clearly indicates that pedestrians are aware of higher chances of getting infected when violating social distancing, and are thus more cautious when walking towards each other. They deliberately enlarge the distance between each other or even try to change the direction, which dramatically decreases the percentage of face-to-face violations from 42.3\% to 20.7\%. 

\begin{figure}[!t]
\begin{center}

\includegraphics[width=0.99\linewidth]{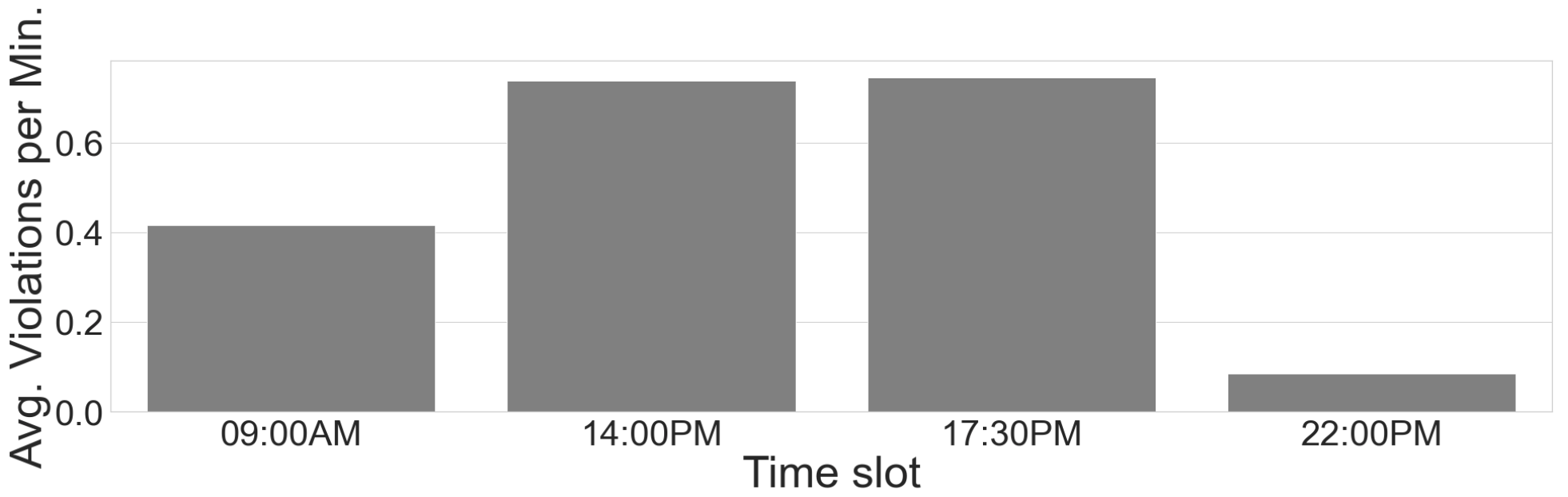}
\vspace*{-0.5cm}
\end{center}
\caption{The number of people who violate social distancing at different times of day.}
\label{nums_people_breaking_time_slot}
\vspace*{-0.2cm}
\end{figure}

Understanding the variability in violations as a function of different times of the day is important. We use a histogram to visualize the statistics of average per minute violations, at different times of day, in Fig.~\ref{nums_people_breaking_time_slot}. The figure shows that there are more than $0.7$ violations per minute during day time (8:00 a.m. to 20:00 p.m.) and less than $0.1$ violations per minute during night time (20:00 p.m. to 08:00 a.m.). Considering that people are more likely to come into contact with each other when crowd density is high, it makes sense that the average number of violations is higher during the day time.

\begin{figure}[!t]
\begin{center}
\includegraphics[width=0.99\linewidth]{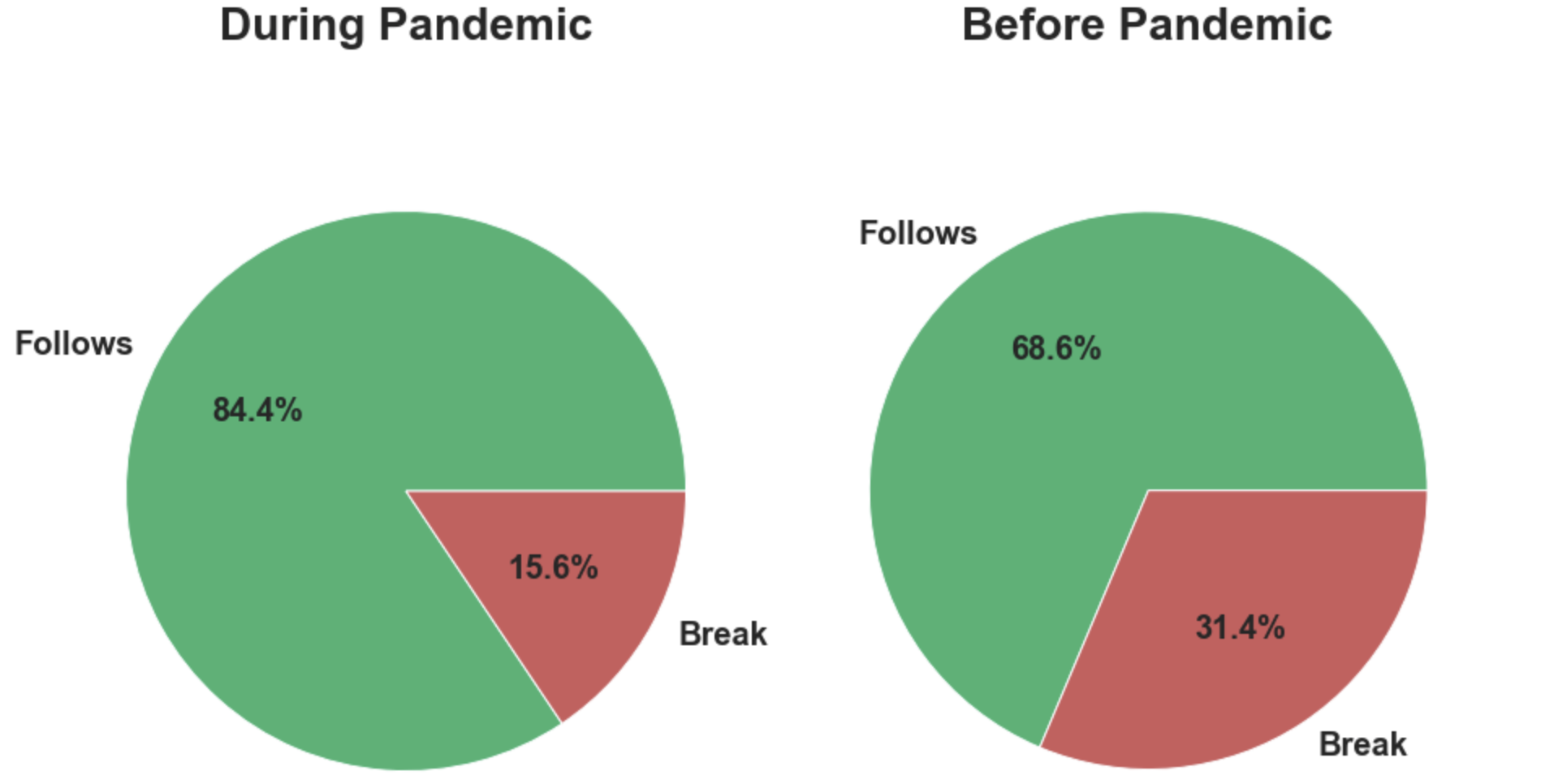}
\vspace*{-0.5cm}
\end{center}
\caption{Pedestrian behavior before and after the social distancing protocol was announced, showing the percentages of rule following and rule breaking occurrences. Note that all rule breaking incidents shorter than 1 second are excluded from this graph. }
\label{old_new_compare}
\end{figure}

Fig.~\ref{old_new_compare} shows the comparison of the proportion of social distancing violations between 2020 (during the pandemic) and 2019 (before the pandemic). One can observe that if we measured the last year’s data using current social distancing rules, 31.4\% of people would be judged as rule violators. However, using the same criteria, only 15.6\% of people are judged as rule violators during the pandemic in 2020. This means that during the pandemic people are aware of social distancing rules, and that they deliberately keep distance from each other. Additionally, home quarantine orders reduce the density of pedestrians on the streets, and the reduced density provides more space for pedestrians to deliberately perform social distancing.

\subsection{The Speed of Inference}
\begin{table}[!t]
\caption{Inference Speed}
\begin{center}
{
\footnotesize

\begin{tabular}{|c|c|}
\hline
 \textbf{Module} & \textbf{Average Frames Per Second (FPS)} \\
\hline\hline
\cline{1-2}
    Detection (YOLOv4 FP32) & 11.07   \\
\cline{1-2}
    Detection (YOLOv4 FP16) & 30.90   \\
\cline{1-2} 
   Tracking (SORT) & 553.83   \\
 \cline{1-2} 
   Social Distancing Analysis & 877.78 \\
\cline{1-2} 
   Overall System (FP32) & 10.72   \\
\cline{1-2} 
   Overall System (FP16) & 28.32   \\
\hline
\end{tabular}
}
\label{inference_speed}
\vspace*{-0.5cm}
\end{center}
\end{table}
\textbf{}
The inference speed of individual model components is shown in Table~\ref{inference_speed}. Note that FP16 TensorRT implementation of YOLOv4 is three times faster than FP32 implementation, without observable loss in object detection accuracy between the two quantization settings.   

\section{Conclusion}

We developed B-SDA, a privacy-preserving system for measurement and analysis of social distancing behavior in traffic intersections based on bird's-eye view video recordings. 
The system relies on customized deep-learning based object detection and tracking techniques, modified and retrained to be able to detect very small pedestrians. 
We introduced a group validation technique to eliminate false positives in detecting social distancing violations, which takes into account movement characteristics including speed, orientation and trajectory to evaluate the  social distancing behavior.
We collected, conditioned and annotated several hundred videos of an intersection in New York City before and during the COVID-19 pandemic. The videos were processed and social distancing analysis was performed. The results were presented in form of probability  functions which capture the type and density of social distancing violations at different times of day and a variety of conditions. Quantitative results correlate well with anticipated and visually observed pedestrian behavior, most dramatically represented by the fact that social distancing violations are estimated to be much less frequent during the pandemic ($15.6\%$) when compared to pre-pandemic statistics ($31.4\%$). 
The analysis indicates that the proposed technique based on bird's-eye view videos is a credible candidate for analysis of social distancing, which could become a part of a real-time epidemiological tool for managing respiratory pandemics. A particular strength of the proposed technique is the privacy-preserving feature, due to the small of pedestrians in bird's-eye view videos. This provides additional motivation to innovate in the area of small object detection and constrained trajectory tracking.

{\small
\bibliographystyle{ieee}
\bibliography{main}

\begin{thebibliography}{10}\itemsep=-1pt

\bibitem{Oxford}
B.~{Benfold} and I.~{Reid}.
\newblock Stable multi-target tracking in real-time surveillance video.
\newblock In {\em CVPR 2011}, pages 3457--3464, 2011.

\bibitem{2019arXiv190305625B}
P.~{Bergmann}, T.~{Meinhardt}, and L.~{Leal-Taixe}.
\newblock {Tracking without bells and whistles}.
\newblock {\em arXiv e-prints}, page arXiv:1903.05625, Mar. 2019.

\bibitem{Bernardin2008Evaluating}
K.~Bernardin and R.~Stiefelhagen.
\newblock Evaluating multiple object tracking performance: the clear mot
  metrics.
\newblock {\em EURASIP Journal on Image and Video Processing}, 2008:1--10,
  2008.

\bibitem{Bewley_2016}
A.~Bewley, Z.~Ge, L.~Ott, F.~Ramos, and B.~Upcroft.
\newblock Simple online and realtime tracking.
\newblock {\em 2016 IEEE International Conference on Image Processing (ICIP)},
  Sep 2016.

\bibitem{2020arXiv200410934B}
A.~{Bochkovskiy}, C.-Y. {Wang}, and H.-Y.~M. {Liao}.
\newblock {YOLOv4: Optimal Speed and Accuracy of Object Detection}.
\newblock {\em arXiv e-prints}, page arXiv:2004.10934, Apr. 2020.

\bibitem{10.1016/j.patcog.2016.06.016}
R.~Chaker, Z.~A. Aghbari, and I.~N. Junejo.
\newblock Social network model for crowd anomaly detection and localization.
\newblock {\em Pattern Recogn.}, 61(C):266–281, Jan. 2017.

\bibitem{VideoAveraging}
C.~COW.
\newblock Averaging video frames, 2012 (accessed July 14, 2020).

\bibitem{cristani2020visual}
M.~Cristani, A.~Del~Bue, V.~Murino, F.~Setti, and A.~Vinciarelli.
\newblock The visual social distancing problem.
\newblock {\em IEEE Access}, 8:126876--126886, 2020.

\bibitem{deepak}
deepak.
\newblock Social-distancing-ai.
\newblock
  \url{https://github.com/deepak112/Social-Distancing-AI/tree/08a9a21ccf8ced3e6ff270628cb1c9b21a55fbee},
  2020.
\newblock Online; accessed 13 July 2020.

\bibitem{2020arXiv200702632E}
M.~{Ehsanpour}, A.~{Abedin}, F.~{Saleh}, J.~{Shi}, I.~{Reid}, and
  H.~{Rezatofighi}.
\newblock {Joint Learning of Social Groups, Individuals Action and Sub-group
  Activities in Videos}.
\newblock {\em arXiv e-prints}, page arXiv:2007.02632, July 2020.

\bibitem{CDC}
C.~for Disease~Control and Prevention.
\newblock Public health guidance for community-related exposure.
\newblock
  \url{https://www.cdc.gov/coronavirus/2019-ncov/php/public-health-recommendations.html},
  2020.
\newblock Online; accessed 31 July 2020.

\bibitem{CDC-social}
C.~for Disease~Control and Prevention.
\newblock Social distancing.
\newblock
  \url{https://www.cdc.gov/coronavirus/2019-ncov/prevent-getting-sick/social-distancing.html},
  2020.
\newblock Online; accessed 31 July 2020.

\bibitem{10.1109/TPAMI.2011.176}
W.~Ge, R.~T. Collins, and R.~B. Ruback.
\newblock Vision-based analysis of small groups in pedestrian crowds.
\newblock {\em IEEE Trans. Pattern Anal. Mach. Intell.}, 34(5):1003–1016, May
  2012.

\bibitem{Girshick_2015_ICCV}
R.~Girshick.
\newblock Fast r-cnn.
\newblock In {\em Proceedings of the IEEE International Conference on Computer
  Vision (ICCV)}, December 2015.

\bibitem{7112511}
R.~{Girshick}, J.~{Donahue}, T.~{Darrell}, and J.~{Malik}.
\newblock Region-based convolutional networks for accurate object detection and
  segmentation.
\newblock {\em IEEE Transactions on Pattern Analysis and Machine Intelligence},
  38(1):142--158, 2016.

\bibitem{gupta2020sd}
S.~Gupta, R.~Kapil, G.~Kanahasabai, S.~S. Joshi, and A.~S. Joshi.
\newblock Sd-measure: A social distancing detector.
\newblock In {\em 2020 12th International Conference on Computational
  Intelligence and Communication Networks (CICN)}, pages 306--311. IEEE, 2020.

\bibitem{Hall1966Hidden}
E.~Hall, E.~Hall, and C.~P. C.~L. of~Congress).
\newblock {\em The Hidden Dimension}.
\newblock Anchor books. Doubleday, 1966.

\bibitem{2017arXiv170306870H}
K.~He, G.~Gkioxari, P.~Dollar, and R.~Girshick.
\newblock Mask r-cnn.
\newblock In {\em Proceedings of the IEEE International Conference on Computer
  Vision (ICCV)}, Oct 2017.

\bibitem{LandingAI}
LandingAI.
\newblock Landing ai creates an ai tool to help customers monitor social
  distancing in the workplace.
\newblock
  \url{https://landing.ai/landing-ai-creates-an-ai-tool-to-help-customers-monitor-social-distancing-in-the-workplace},
  2020.
\newblock Online; accessed 13 July 2020.

\bibitem{2018arXiv181111968L}
N.~{Liu}, Y.~{Long}, C.~{Zou}, Q.~{Niu}, L.~{Pan}, and H.~{Wu}.
\newblock {ADCrowdNet: An Attention-injective Deformable Convolutional Network
  for Crowd Understanding}.
\newblock {\em arXiv e-prints}, page arXiv:1811.11968, Nov. 2018.

\bibitem{liu2016ssd}
W.~Liu, D.~Anguelov, D.~Erhan, C.~Szegedy, S.~Reed, C.-Y. Fu, and A.~C. Berg.
\newblock Ssd: Single shot multibox detector.
\newblock In {\em European conference on computer vision}, pages 21--37.
  Springer, 2016.

\bibitem{8953871}
Y.~{Liu}, M.~{Shi}, Q.~{Zhao}, and X.~{Wang}.
\newblock Point in, box out: Beyond counting persons in crowds.
\newblock In {\em 2019 IEEE/CVF Conference on Computer Vision and Pattern
  Recognition (CVPR)}, pages 6462--6471, 2019.

\bibitem{2014arXiv1409.7618L}
W.~Luo, J.~Xing, A.~Milan, X.~Zhang, W.~Liu, and T.~Kim.
\newblock Multiple object tracking: A literature review.
\newblock {\em Artificial Intelligence}, 293:103448, 2014.

\bibitem{magdy2015review}
N.~Magdy, M.~A. Sakr, T.~Mostafa, and K.~El-Bahnasy.
\newblock Review on trajectory similarity measures.
\newblock In {\em 2015 IEEE seventh international conference on Intelligent
  Computing and Information Systems (ICICIS)}, pages 613--619. IEEE, 2015.

\bibitem{Proof_COVID}
L.~Matrajt and T.~Leung.
\newblock Evaluating the effectiveness of social distancing interventions
  against covid-19.
\newblock {\em medRxiv}, 2020.

\bibitem{punn2020monitoring}
N.~S. Punn, S.~K. Sonbhadra, and S.~Agarwal.
\newblock Monitoring covid-19 social distancing with person detection and
  tracking via fine-tuned yolo v3 and deepsort techniques.
\newblock {\em arXiv preprint arXiv:2005.01385}, 2020.

\bibitem{redmon2016you}
J.~Redmon, S.~Divvala, R.~Girshick, and A.~Farhadi.
\newblock You only look once: Unified, real-time object detection.
\newblock In {\em Proceedings of the IEEE conference on computer vision and
  pattern recognition}, pages 779--788, 2016.

\bibitem{2016arXiv161208242R}
J.~{Redmon} and A.~{Farhadi}.
\newblock Yolo9000: Better, faster, stronger.
\newblock In {\em 2017 IEEE Conference on Computer Vision and Pattern
  Recognition (CVPR)}, pages 6517--6525, 2017.

\bibitem{2018arXiv180402767R}
J.~Redmon and A.~Farhadi.
\newblock Yolov3: An incremental improvement.
\newblock {\em Computer Vision and Pattern Recognition}, 2018.

\bibitem{2015arXiv150601497R}
S.~{Ren}, K.~{He}, R.~{Girshick}, and J.~{Sun}.
\newblock Faster r-cnn: Towards real-time object detection with region proposal
  networks.
\newblock {\em IEEE Transactions on Pattern Analysis and Machine Intelligence},
  39(6):1137--1149, 2017.

\bibitem{rezaei2020deepsocial}
M.~Rezaei and M.~Azarmi.
\newblock Deepsocial: Social distancing monitoring and infection risk
  assessment in covid-19 pandemic.
\newblock {\em Applied Sciences}, 10(21):7514, 2020.

\bibitem{2014arXiv1409.0575R}
O.~{Russakovsky}, J.~{Deng}, H.~{Su}, J.~{Krause}, S.~{Satheesh}, S.~{Ma},
  Z.~{Huang}, A.~{Karpathy}, A.~{Khosla}, M.~{Bernstein}, A.~C. {Berg}, and
  L.~{Fei-Fei}.
\newblock {ImageNet Large Scale Visual Recognition Challenge}.
\newblock {\em arXiv e-prints}, page arXiv:1409.0575, Sept. 2014.

\bibitem{2020arXiv200501385S}
N.~{Singh Punn}, S.~K. {Sonbhadra}, and S.~{Agarwal}.
\newblock {Monitoring COVID-19 social distancing with person detection and
  tracking via fine-tuned YOLO v3 and Deepsort techniques}.
\newblock {\em arXiv e-prints}, page arXiv:2005.01385, May 2020.

\bibitem{6636608}
F.~{Solera}, S.~{Calderara}, and R.~{Cucchiara}.
\newblock Structured learning for detection of social groups in crowd.
\newblock In {\em 2013 10th IEEE International Conference on Advanced Video and
  Signal Based Surveillance}, pages 7--12, 2013.

\bibitem{sun2020autonomous}
P.~Sun, G.~Draughon, and J.~Lynch.
\newblock An autonomous approach to measure social distances and hygienic
  practices during covid-19 pandemic in public open spaces.
\newblock {\em arXiv preprint arXiv:2011.07375}, 2020.

\bibitem{2018arXiv181011780S}
S.~{Sun}, N.~{Akhtar}, H.~{Song}, A.~{Mian}, and M.~{Shah}.
\newblock Deep affinity network for multiple object tracking.
\newblock {\em IEEE Transactions on Pattern Analysis and Machine Intelligence},
  43(1):104--119, 2018.

\bibitem{wojke2017simple}
N.~Wojke, A.~Bewley, and D.~Paulus.
\newblock Simple online and realtime tracking with a deep association metric.
\newblock In {\em 2017 IEEE international conference on image processing
  (ICIP)}, pages 3645--3649. IEEE, 2017.

\bibitem{2020arXiv200703578Y}
D.~{Yang}, E.~{Yurtsever}, V.~{Renganathan}, K.~A. {Redmill}, and
  {\"U}.~{{\"O}zg{\"u}ner}.
\newblock {A Vision-based Social Distancing and Critical Density Detection
  System for COVID-19}.
\newblock {\em arXiv e-prints}, page arXiv:2007.03578, July 2020.

\bibitem{2019arXiv191108287Z}
Z.~{Zheng}, P.~{Wang}, W.~{Liu}, J.~{Li}, R.~{Ye}, and D.~{Ren}.
\newblock {Distance-IoU Loss: Faster and Better Learning for Bounding Box
  Regression}.
\newblock {\em arXiv e-prints}, page arXiv:1911.08287, Nov. 2019.

\bibitem{2020arXiv200106303Z}
P.~{Zhu}, L.~{Wen}, D.~{Du}, X.~{Bian}, Q.~{Hu}, and H.~{Ling}.
\newblock {Vision Meets Drones: Past, Present and Future}.
\newblock {\em arXiv e-prints}, page arXiv:2001.06303, Jan. 2020.

\end{thebibliography}
}

\end{document}